\documentclass[letterpaper, 10 pt, journal]{ieeeconf}
\usepackage[english]{babel}
\IEEEoverridecommandlockouts                              

\usepackage{amsmath}
\usepackage{graphicx}
\usepackage{verbatim}
\usepackage[colorlinks=true, allcolors=blue]{hyperref}
\usepackage{multirow}
\usepackage{amsfonts}
\usepackage{siunitx}
\usepackage{subcaption}

\title{Map Prediction and Generative Entropy for Multi-Agent Exploration}
\author{Alexander Spinos$^{1}$, Bradley Woosley$^{2}$, Justin Rokisky$^{1}$, Christopher Korpela$^{1}$, \\ John G. Rogers III$^{2}$, Brian A. Bittner$^{1}$
\thanks{$^{1}$Johns Hopkins University Applied Physics Lab, Laurel, MD, USA.
\newline \indent $^{2}$DEVCOM Army Research Lab, Adelphi, MD, USA.    
\newline \indent Distribution statement A. Approved for public release; distribution is unlimited
       }
}
\begin{document}
\maketitle

\begin{abstract}
Traditionally, autonomous reconnaissance applications have acted on explicit sets of historical observations. Aided by recent breakthroughs in generative technologies, this work enables robot teams to act beyond what is currently known about the environment by inferring a distribution of reasonable interpretations of the scene. We developed a map predictor that inpaints the unknown space in a multi-agent 2D occupancy map during an exploration mission. From a comparison of several inpainting methods, we found that a fine-tuned latent diffusion inpainting model could provide rich and coherent interpretations of simulated urban environments with relatively little computation time. By iteratively inferring interpretations of the scene throughout an exploration run, we are able to identify areas that exhibit high uncertainty in the prediction, which we formalize with the concept of generative entropy. We prioritize tasks in regions of high generative entropy, hypothesizing that this will expedite convergence on an accurate predicted map of the scene. In our study we juxtapose this new paradigm of task ranking with the state of the art, which ranks regions to explore by those which maximize expected information recovery. We compare both of these methods in a simulated urban environment with three vehicles. Our results demonstrate that by using our new task ranking method, we can predict a correct scene significantly faster than with a traditional information-guided method.
\end{abstract}

\vspace{-3mm}

\section{Introduction}
\begin{figure}
\centering
\includegraphics[width=0.9\columnwidth]{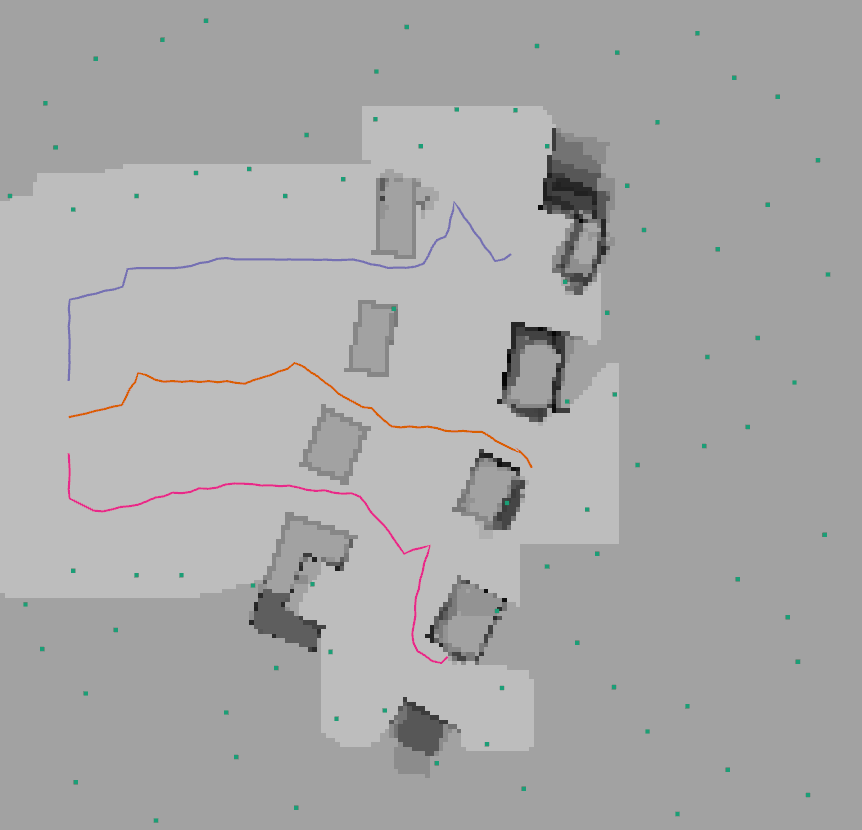}
\caption{Visualization of the generative entropy overlaid on the observed map during an exploration trial with three agents (paths shown in color). Dark regions indicate areas of high generative entropy, which are prioritized to resolve the remaining uncertainty in the predicted map. Task locations are shown as green squares.
}
\label{fig:map_prediction_exploration_run2}
\end{figure}

At a high level, the goal of multi-robot exploration is to grow and refine an environmental representation of an area by sending robots to locations or poses where they can observe currently unknown space.
The challenge is to efficiently prioritize and delegate these navigation tasks among the agents to improve the resulting map quality or reduce the time to completion.
The field of multi-agent exploration has advanced greatly in maturity, with several universities demonstrating robust approaches to mapping and asset localization in unstructured cave-like environments \cite{zhao2021super,tranzatto2022cerberus}. 
Occupancy grids are the traditional choice for environmental representation, but new approaches using semantic meshes and 3D scene graphs have emerged with significant advancements in both the quality and density of information that can be shared across agents \cite{chang2023hydra,gu2024conceptgraphs}. 
Current challenges exist in how to synthesize this information into guidance that can be leveraged for higher efficiency in exploration missions.

For efficient task delegation, the field of multi-robot exploration includes a collection of task allocation methods that prioritize tasks based on the currently known state of the environment.
These methods span a variety of communication and computation architectures, with more robust methods targeting distributed systems with unreliable communication \cite{nayak2020experimental, woosley2021bid}.
The state-of-the-art reward function for 2D and 3D occupancy grid environments is \textit{mutual information}, which captures the expected information gain at a location given the robot sensor model \cite{julian2014mutual, charrow2015information, henderson2020efficient}.
By assigning tasks to maximize the mutual information recovered over some time window, a team of robots can uncover more absolute map information in a shorter amount of time.

In parallel, rapid advancements in the maturity of generative technologies allow us to repeatably generate realistic scenes (e.g. as images) from little-to-no data \cite{rombach2022high,lugmayr2022repaint,sun2024autoregressive}.
In this work, we show that we can use such methods to predict a likely state of the entire environment based on our current observations.
This predicted environment can then be used to prioritize exploration tasks with greater foresight, reducing the time required to produce a high-quality map of a specified area.
In particular, we will predict 2D occupancy grid environments using inpainting methods.
Diffusion-based inpainting models are particularly well suited to the multi-robot mapping case, as they can synthesize a complete scene from disparate observations across the map made by any number of agents.

Prior work has shown how environment prediction can improve robot performance, particularly for navigation tasks.
Inpainting methods and related prediction architectures have been used to extend occupancy maps generated by limited field of view sensors to improve navigation efficiency \cite{wei2021occupancy, sharma2023proxmap}.
Generative Adversarial Networks (GANs) have also been used to predict local maps for high speed navigation around corners \cite{katyal2021high}.
However, this result relied on the local distribution inference capabilities of GANs \cite{bau2019seeing} and did not extend well to predicting more global scenes.
In the robot exploration domain, Shrestha et. al. showed that using a Variational Autoencoder to predict the map in the area around frontier points can improve the accuracy of the expected information gain metric, boosting exploration efficiency \cite{shrestha2019learned}.
In general, these prior works focus on training relatively small models to predict a limited area around the robot or around frontiers, seeking to produce stable and reliable output to inform navigation or information recovery.

In contrast, we seek to predict large-scale features throughout the global environment by fine-tuning larger, pre-trained inpainting models.
Fine tuning allows us to leverage any existing ability to inpaint birds-eye images of buildings, which we expect to be a meaningful component of the pre-trained models. 
In future work, we can also build on any existing capability for these models to inpaint on rich sets of semantic labels.
When the known information is limited (for example, at the beginning of exploration), attempting to inpaint unknown regions will yield high variation in the predicted output. 
Rather than being a detriment to performance, this high variation is a valuable indicator of regions which are ``hard to predict,'' and thus more important to prioritize during exploration.
We formalize this notion with the concept of \textit{generative entropy}, which we use as a metric for task allocation.

Traditionally, mapping approaches that use mutual information proceed on the assumption that the information contained in every map cell is independent and equally valuable.
However, this assumption is often faulty and counterproductive.
Consider a small town next to an adjacent open field.
After a few measurements, the robot may see that the town is full of obstacles, while the open field is relatively free of obstacles.
In a system driven by mutual information, the robot would prefer to explore the open field, where the laser sensor can easily uncover more cells compared to the cluttered town.
However, the town is clearly the more interesting and ``information-rich'' region for any practical application.
Within our map prediction framework, the contents of the open field are more easily predicted, while the town would have higher generative entropy and attract the robots.
Figure~\ref{fig:map_prediction_exploration_run2} shows an example of a town exploration task with an overlaid generative entropy field; regions of high entropy are clustered around building edges that need to be observed to resolve the map geometry.

Once the critical sections of the map necessary to understand the scene have been explored, the predicted map provides a reliable estimate of the final map.
This estimate can be produced much sooner than it would take to explicitly observe the entire map, and can provide key situational awareness to a user of the system in situations where time is of the essence, such as search-and-rescue scenarios.
In this paper, we will demonstrate this benefit through simulated experiments using both traditional task allocation and our proposed generative entropy method, finding that the latter approach pushes this advantage even further.

This framework has several additional desirable properties.
The runtime of the diffusion model is independent of the number of agents, and is thus suitable for robot teams of any size.
The predictive output is used only for task weighting and is not directly used for navigation, so there is no impact to robot safety in the event of hallucinations or mispredictions.
While the results in this paper are limited in scope, we believe this framework can be extended to more sophisticated environmental representations, and would improve in both generality and predictive power if allowed to ingest semantic information as well as geometric data.

\section{Problem Definition and Approach}
\label{section:problem_definition}

In this work, $n$ agents are assigned to explore a specified, bounded area for a reconnaissance mission.
The environments are ``urban'' outposts or small towns, characterized by large open spaces and a small number of buildings.
The goal is to produce a useful map of the specified area as soon as possible.

As the robots explore, each individual robot produces a local occupancy map from lidar observations, which are fused online to update a global occupancy map.
This global map of observations is periodically used to predict a completed global map using the map prediction model, described in detail in Section~\ref{section:map_prediction}.
The stream of predicted maps is used to calculate a generative entropy field, described in Section~\ref{section:generative_entropy}, which is used by the task allocator to weight the importance of different task locations.

Since the area to explore is bounded, task locations can either be extracted as frontier points on the border of known and unknown space, or they can be distributed evenly throughout the environment.
We use a combination of the two approaches. 
Initially, quasirandom points are scattered over the entire environment to allow the task allocator to evaluate the importance of different regions and efficiently route robots over a long horizon.
During exploration, a frontier extractor may add additional tasks where the space is not adequately covered to ensure map completeness.

Tasks are allocated among the robots in a distributed manner using a market-based algorithm, ACBBA \cite{johnson2010improving}.
In this work, the map fusion, map prediction, and frontier extraction are performed in a centralized manner for simplicity. However, it would be relatively straightforward to replicate these calculations on each robot to enable a fully distributed system.
Experiments are performed in a simulated multi-agent environment, described in detail in Section~\ref{section:exploration_performance}.

\section{Multi-Agent Map Prediction}
\label{section:map_prediction}

Here we will take an occupancy grid map that is shared amongst a number of agents (three in this work). In our simulated analyses this map will be a perfect fusion of local maps. 
We will seek to generate an estimated map from a partially observed map, which may contain largely unobserved data. We generate this estimated map by transforming the problem into an image inpainting \cite{xiang2023deep} problem and leveraging recent advancements in this domain \cite{rombach2022high,lugmayr2022repaint,sun2024autoregressive}. Each map (200 rows by 200 columns) at 0.5 meters per pixel is converted to a quantized grayscale image and a binary mask. To transform the occupancy map into an image, first values from $(0,1)$ in the occupancy space are mapped to discrete values of known occupied, known free, and unknown which are then mapped to colors black, white, and gray respectively in the grayscale pixel space. A binary inpainting mask is then generated from the unknown/gray pixels. Both grayscale and mask images are resized to either 256x256 or 512x512 depending on the model. The grayscale occupancy map and the binary mask are fed into an inpainting model, fine-tuned for this task, which completes the unknown region as classified by the binary mask. The output image is then converted back into an occupancy map and provided to the rest of the system for identifying areas of maximum interest.

\subsection{Models Tested}

After reviewing the state of the art in image inpainting, we selected three models to evaluate for use in a multi-robot exploration architecture. Here we explain each model and highlight distinguishing strengths and weaknesses. 

LaMa, Large Mask inpainting \cite{suvorov2021resolution}, pairs the large receptive field of Fast Fourier Convolutions (FFC) \cite{NEURIPS2020_2fd5d41e} with a high receptive field  perceptual loss and a dynamic large mask generation strategy to inpaint the masked area. For this work we fine-tuned the Big-LaMa model to a training set we discuss in \ref{subsection:training_set}. 
The inability to condition on semantic information is the primary shortcoming for long term use of this prediction model within broader exploration architectures, since we anticipate the desire to incorporate semantic knowledge into scene prediction.

RePaint \cite{lugmayr2022repaint} resamples a fine-tuned unconditional Denoising Diffusion Probabilistic Model (DDPM) \cite{ho2020denoisingdiffusionprobabilisticmodels} while conditioning the unknown region of the image on the known region to inpaint a masked area. As no masks are used in the training process, this approach is more robust to unseen mask geometries. These pixel-space diffusion steps are computationally costly, but embed high complexity structure in the denoiser. For this work we fine-tuned an unconditional DDPM for 100 epochs and ran the RePaint pipeline using the HuggingFace Diffuser library \cite{von_Platen_Diffusers_State-of-the-art_diffusion}. 

Stable Diffusion \cite{rombach2022high} learns a denoising model that acts in a latent image space. While it cannot encode the complexity of denoisers available in the pixel diffusion model, it can run at a much higher rate. The latent space has been optimized to contain a highly relevant embedding of the image space (with respect to the large dataset of images it is trained on). This embedding could be further curated and specialized for the space of occupancy map images and could be the subject of future work. We fine-tuned the Stable Diffusion v1.2 inpainting model using a fixed instance prompt of ``an occupancy grid'' for 250,000 steps, again using the HuggingFace Diffuser library \cite{von_Platen_Diffusers_State-of-the-art_diffusion}.

\subsection{Notes on Masking}
\begin{figure}
\centering
\includegraphics[width=1\columnwidth, trim={3cm 0 2cm 0}, clip]{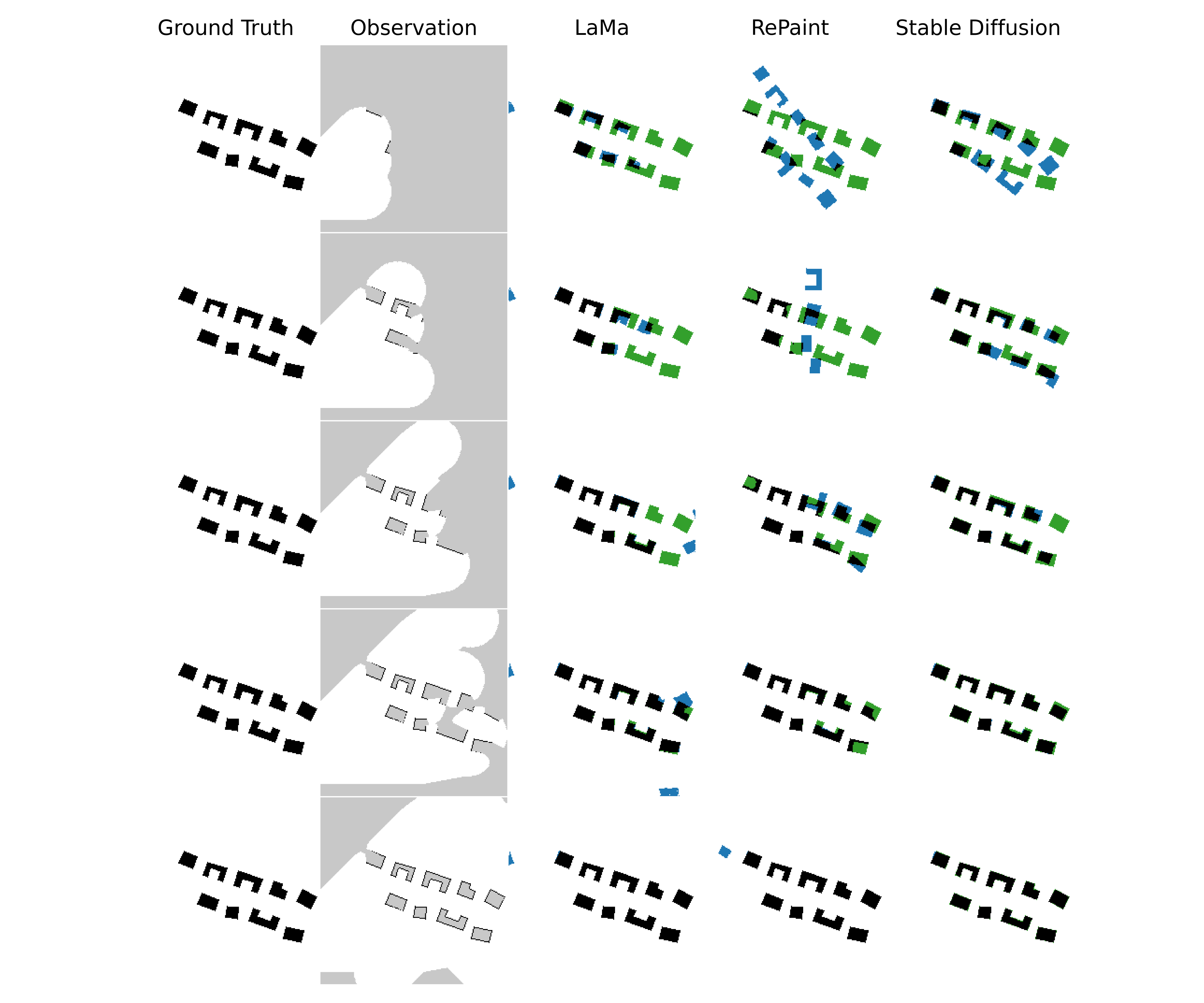}
\caption{Map predictions for each model when 20\%, 35\%, 50\%, 65\%, and 80\% of the map information was observed during a single trial. Masks of complex geometry (grey, second column) are inpainted by LaMa, Repaint, and Stable Diffusion. Performance is presented (last three columns) as correctly predicted free (white), correctly predicted occupied (black), incorrectly predicted free (green), and incorrectly predicted occupied (blue).
}
\label{fig:prediction_comparison}
\end{figure}

LaMa and Stable Diffusion both require masks as inputs during the training process. We first fine-tuned and evaluated both models using their default masking process. These involved dynamically generated diverse large masks for LaMa and a single dynamically generated rectangle and ellipse for Stable Diffusion. We found the robust masking process of LaMa coupled with their model architecture resulted in strong performance when given the out-of-domain masks of our map prediction problem, which involve the geometric complexity of scene observation data, as can be seen in Figure \ref{fig:prediction_comparison}. The simpler masking process of Stable Diffusion resulted in the model struggling with the out-of-domain data. To remedy this, we used map exploration observation data captured at a variety of timestamps when generating the inpainting training set (described in \ref{subsection:training_set}). 
Examples can be seen in the second column in Figure \ref{fig:prediction_comparison}; the gray pixels which correspond to unknown space are converted to inpainting masks during training. Using in-domain masks greatly improved the performance of Stable Diffusion. We also used these observation masks while fine-tuning LaMa without a significant change in performance. 


\begin{table*}
\centering
\begin{tabular}{c c c c c c c}
\hfill\\[-1.5mm]
 \hline
 Model & LPIPS & LPIPS @ 20\% & LPIPS @ 35\% & LPIPS @ 50\% & LPIPS @ 65\% & LPIPS @ 80\%\\
 \hline
 LaMa & 0.095 & 0.197 & 0.118 & 0.079 & 0.053 & 0.027 \\
 RePaint & 0.120 & 0.306 & 0.143 & 0.080 & 0.049 & 0.020\\
 Stable Diffusion & 0.088 & 0.192 & 0.108 & 0.067 & 0.046 & 0.027 \\
 \hline
\end{tabular}
\caption{LPIPS is reported as the multi-agent fleet recovers various percentages of the total map information (top row).}
\label{table:metrics_lpips}
\end{table*}

\begin{table*}
\centering
\begin{tabular}{c c c c c c c} 
 \hline
 Model & FID & FID @ 20\% & FID @ 35\% & FID @ 50\% & FID @ 65\% & FID @ 80\%\\
 \hline
 LaMa & 30.29 & 74.15 & 35.86 & 30.81 & 21.83 & 11.25 \\
 RePaint & 8.70 & 33.87 & 12.92 & 6.56 & 3.59 & 1.36\\
 Stable Diffusion & 28.02 & 27.95 & 28.66 & 29.47 & 29.78 & 30.06 \\
 \hline
\end{tabular}
\caption{FID is reported as the multi-agent fleet recovers various percentages of the total map information (top row).}
\label{table:metrics_fid}
\end{table*}

\subsection{Training Set}
\label{subsection:training_set}
We created a 2-D procedural map generator and multi-robot exploration simulation environment to generate a large quantity of maps and observation data snapshots. 
The map generator produces occupancy grids of simple town environments.
It generates a main street with random orientation and curvature, then selects a random number of buildings to place along the street.
Buildings are drawn from a small set of parameterized types (rectangular, L-shaped, and C-shaped) with randomized shape parameters, and placed on both sides of the street with some randomness in the relative spacing and angular orientation.
The intent with this dataset was not to capture a large variety of possible environments, but rather to see if the model could capture both local features (ability to complete partially observed buildings) and global structure (infer likely locations of unobserved buildings given the alignment of observed buildings).

Given a generated map, we produce several partial observations of this map using a simple robot simulator.
Three robots are initialized in a random location along the edge of the map, with random heading directions biased towards the map center.
The robots are driven to move in their desired heading direction, avoiding obstacles using a potential field method.
If the robots reach a local minimum or hit a map edge, then they set a new random goal direction, continuing to bounce around until the map is fully explored.
As the robots move, they uncover map cells using a simulated lidar model.
Snapshots of the exploration state are taken periodically when certain thresholds of map coverage are passed.
Training data snapshots are taken at 10\% intervals, and test data snapshots are taken at five more coarsely spaced intervals.
An example generated map and test data exploration snapshots are shown in columns 1 and 2 in Figure~\ref{fig:prediction_comparison}.

Multiple exploration rollouts were conducted for each map. 
In total, 2000 maps and 180,000 exploration snapshots were generated for the training set, with 2000 maps and 30,000 snapshots generated for the test set.

\section{Performance of Map Prediction on an Urban Environment Set}

The map generator and simulation described in \ref{subsection:training_set} were used to generate a held-out dataset for evaluation.
While we used the default inference settings for LaMa, we adjusted the number of inference steps for RePaint and Stable Diffusion to balance speed and performance. For RePaint, we used 100 steps with a jump length of 10 and a jump $n$ sample of 10, and for Stable Diffusion, we used 50 steps. Inference took roughly 30 seconds per image for RePaint, roughly 1 second per image for Stable Diffusion, and roughly 0.2 seconds for LaMa when running the models on an Nvidia A100. 
We evaluated two common metrics for inpainting performance: Learned Perceptual Image Patch Similarity (LPIPS) \cite{zhang2018unreasonable}, Table \ref{table:metrics_lpips}, and Fréchet Inception Distance (FID) \cite{heusel2017gans}, Table \ref{table:metrics_fid}. These were evaluated at 5 different levels of map exploration completeness: 20\%, 35\%, 50\%, 65\%, \& 80\%. The intervals were selected due to the general lack of new occupied space observations in the first and last 20\% of information extraction. The LPIPS performance is generally consistent across models, while there is variance in FID results. 

Notably, Stable Diffusion offers better performance at the early stages of exploration whereas RePaint offers better performance near the end of the exploration run. Due to the large difference in prime objective between map prediction and general image inpainting (accurately generating obstacles vs generating realistic image data), we consider that traditional inpainting metrics may not provide a comprehensive look at qualitative assessments. 

To gain an intuitive understanding of relative model performance we provide a visualization to show the accuracy of predictions at multiple observation levels in Figure~\ref{fig:prediction_comparison}. 
In Figure~\ref{fig:prediction_comparison}, we observe occasional outlier buildings on the map periphery for LaMa and an initial misassignment of the ``main road'' for RePaint. While these characteristics appeared to be representative of differentiating factors in performance, a component that we generally observed (that is not observed in Figure~\ref{fig:prediction_comparison}) is the tendency for RePaint to outperform Stable Diffusion towards the end of the exploration trial. We ultimately selected Stable Diffusion as our model of choice due to its shorter run time, high performance on our test dataset, and the ability to potentially condition the inpainting process on other input such as text descriptions, images, etc in future work.


\section{Generative Entropy}
\label{section:generative_entropy}

Here we seek to incorporate map prediction to more rapidly extract high accuracy maps of new environments.
We would like to direct the robots towards regions where additional observations will improve the accuracy of the map predictor, which
we accomplish by defining \emph{generative entropy} as a metric for task weighting.

The environment is modeled as an occupancy grid $m = (m_1, m_2, ...)$ with each cell $m_i \in \{0,1\}$.
Throughout exploration, we maintain a belief about each cell in the environment $p_i = p(m_i)$ which represents a probability that the cell is occupied.
Each time a predicted map is produced, it is treated like a noisy observation and used to update the posterior probability of every cell in the occupancy grid.
Cells which are consistently predicted as free or occupied will increase the confidence in this belief, up to a saturation threshold. 
We define the generative entropy $H_i$ of a map cell as the binary entropy of this estimated occupancy probability:
\begin{equation}
H_i = -p_i log_{2}(p_i) - (1-p_i) log_{2}(1-p_i)
\end{equation}

\begin{figure}
\centering
\hfill\\[1.5mm]
\includegraphics[width=.32\linewidth]{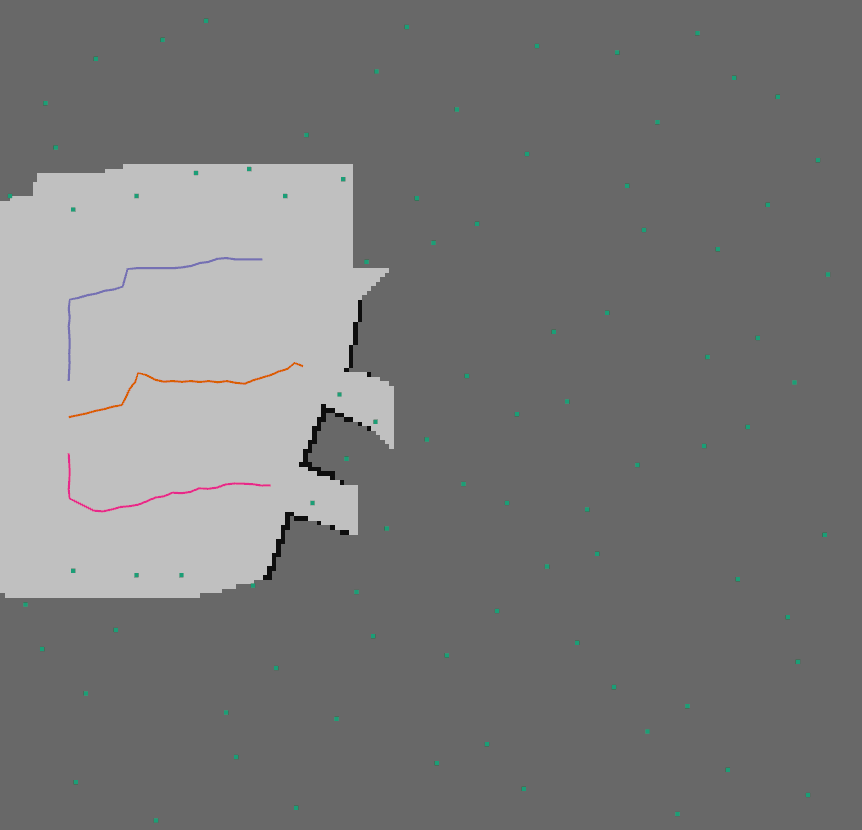}\,%
\includegraphics[width=.32\linewidth]{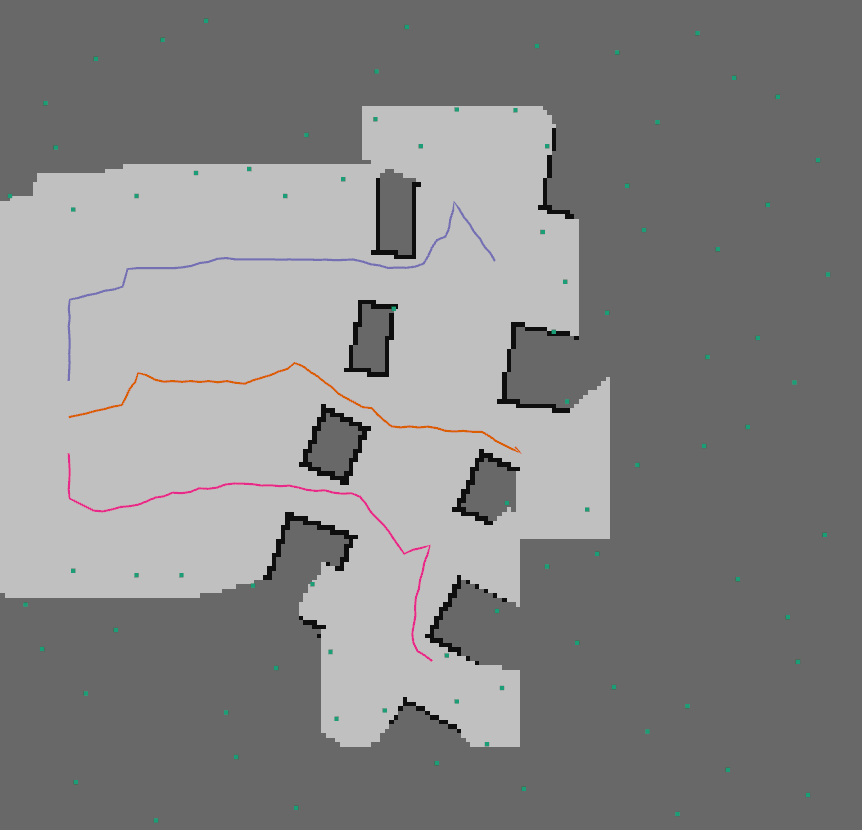}\,%
\includegraphics[width=.32\linewidth]{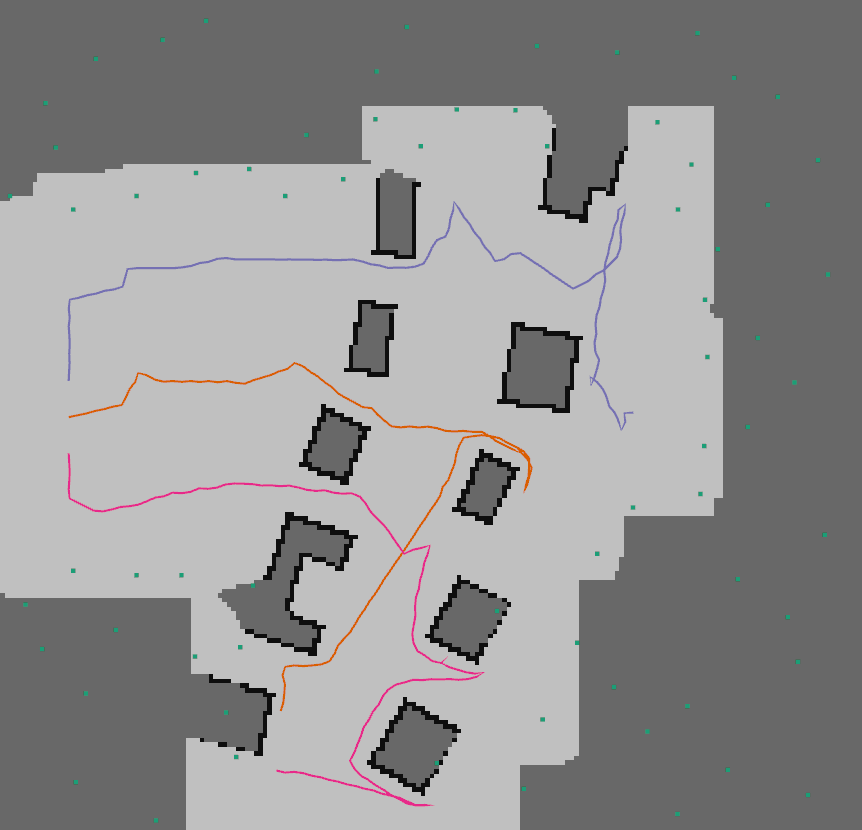}
\\[2pt]
\includegraphics[width=.32\linewidth]{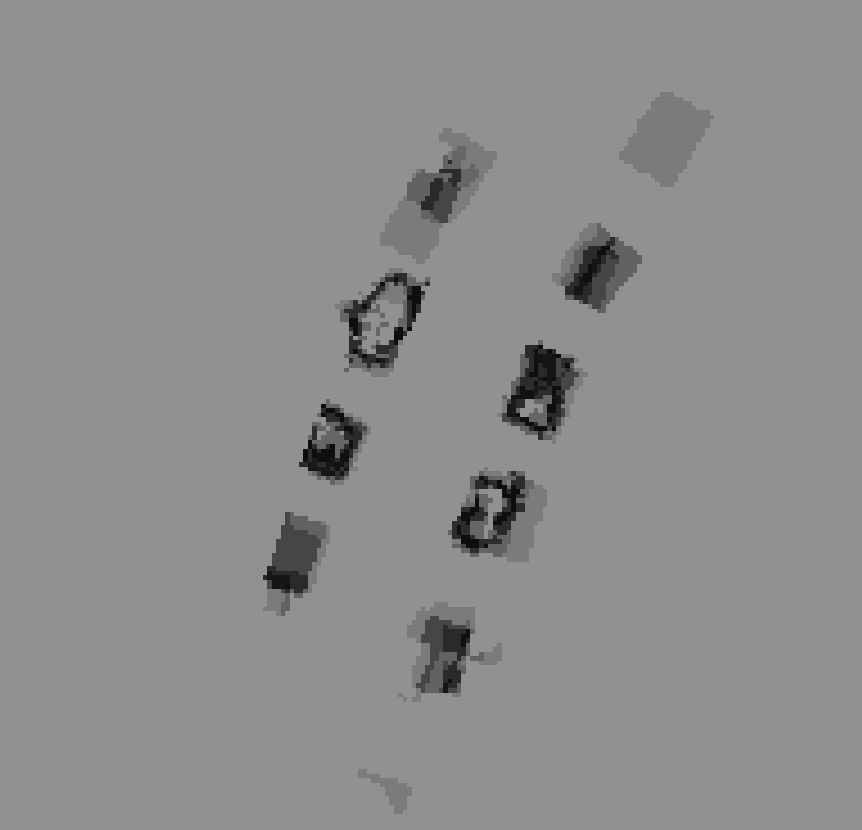}\,%
\includegraphics[width=.32\linewidth]{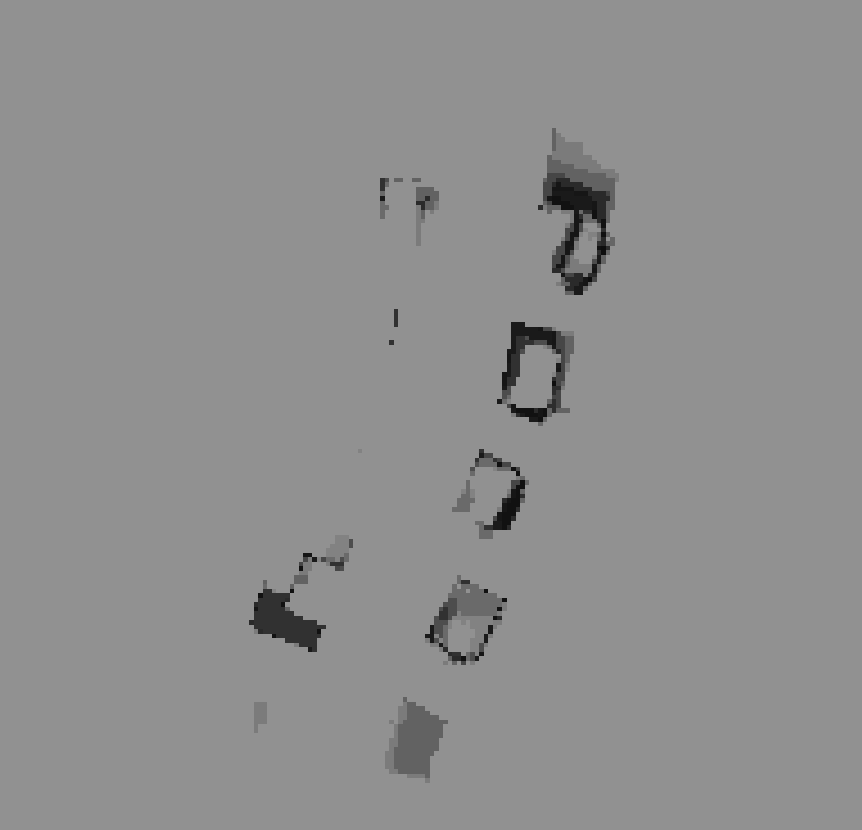}\,%
\includegraphics[width=.32\linewidth]{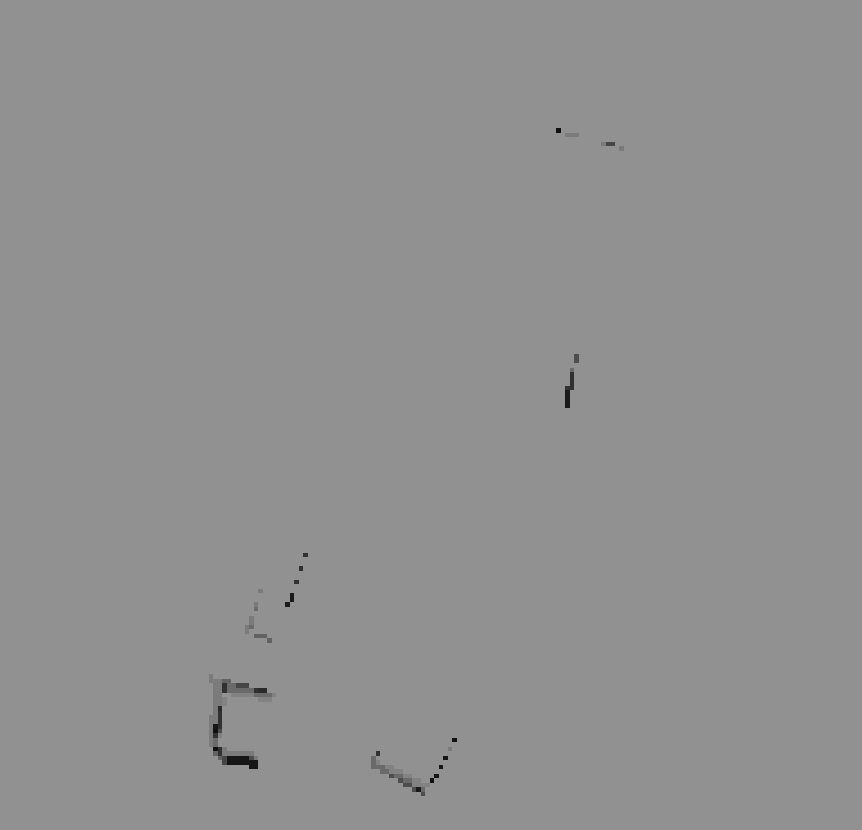}
\caption{Top row: simulated exploration of an urban environment with three agents, with images corresponding to 20\%, 35\%, and 50\% cells observed. Bottom row: the corresponding generative entropy field. Darker pixels correspond to points with high entropy.
}
\label{fig:inpainting_entropy}
\end{figure}

Thus, cells which exhibit large variation in the generative output throughout the course of exploration gain larger generative entropy.
An example of the generative entropy map during a simulated experiment is shown in Figure~\ref{fig:inpainting_entropy}.
At 20\% of cells observed, the map predictor can reliably guess the orientation of the main road, but there are still large regions of uncertainty near suspected building locations.
At 35\% of cells observed, the centers of partially observed buildings are low entropy because the map predictor will reliably attempt to complete a building in that location. The building edges, however, may be high entropy, since the shape of the building will vary over successive predictions.
At 50\% of cells observed, only a few small pockets of high entropy remain.

\section{Simulated Exploration Experiments}
\label{section:exploration_performance}

Here we will discuss the experiments run to determine the advantage of map prediction for exploration.
We created a robot simulator to test the approach described in Section~\ref{section:problem_definition}.
Three robots are tasked to explore an environment generated by the procedural map generator.
We test three different task reward settings to inform the task allocator (ACBBA): \emph{Constant}, \emph{Visible Entropy}, and \emph{Generative Entropy}.
Each setting was tested with 10 trials.

In ACBBA, robots bid to add tasks to their task bundle, scoring tasks based on the time-discounted reward \cite{choi2009consensus}.
For robot $i$ planning to visit a bundle of tasks that form a path $\mathbf{p}_i$, the bundle score is defined as follows:

\begin{equation}
S_i^{\mathbf{p}_i} = \sum_j \lambda^{\tau_i^j (\mathbf{p}_i)} c_j
\end{equation}

where $\lambda < 1$ is a fixed discount factor, $\tau_i^j (\mathbf{p}_i)$ is the expected time to reach task $j$ along the path, and $c_j$ is the pre-discount reward of task $j$.
The reward setting determines the value of $c_j$.
\emph{Constant} gives all tasks a constant pre-discount reward.
\emph{Visible Entropy} is an information-theoretic method, which gives each task a reward scaled with the volume of unknown space the robot would expect to observe from that location.
\emph{Generative Entropy} scales the task reward with the total generative entropy in a box centered on the task point. 

In all trials, we use a bundle size of 3, discount factor of 0.95, sensor radius of \SI{10}{m}, map size of 100x\SI{100}{m}, and a map prediction period of \SI{2.5}{s}.
An example of an in-progress trial with \emph{Generative Entropy} is shown in Figure~\ref{fig:inpainting_entropy}.
Throughout the exploration run, we measure two quantities: total percentage of the map explored in terms of absolute cells uncovered, and accuracy of the latest predicted map compared to the ground truth map.


\subsection{Results}

\begin{figure}
\centering
\includegraphics[width=\columnwidth, trim={1cm 1cm 1.6cm 1cm}, clip]{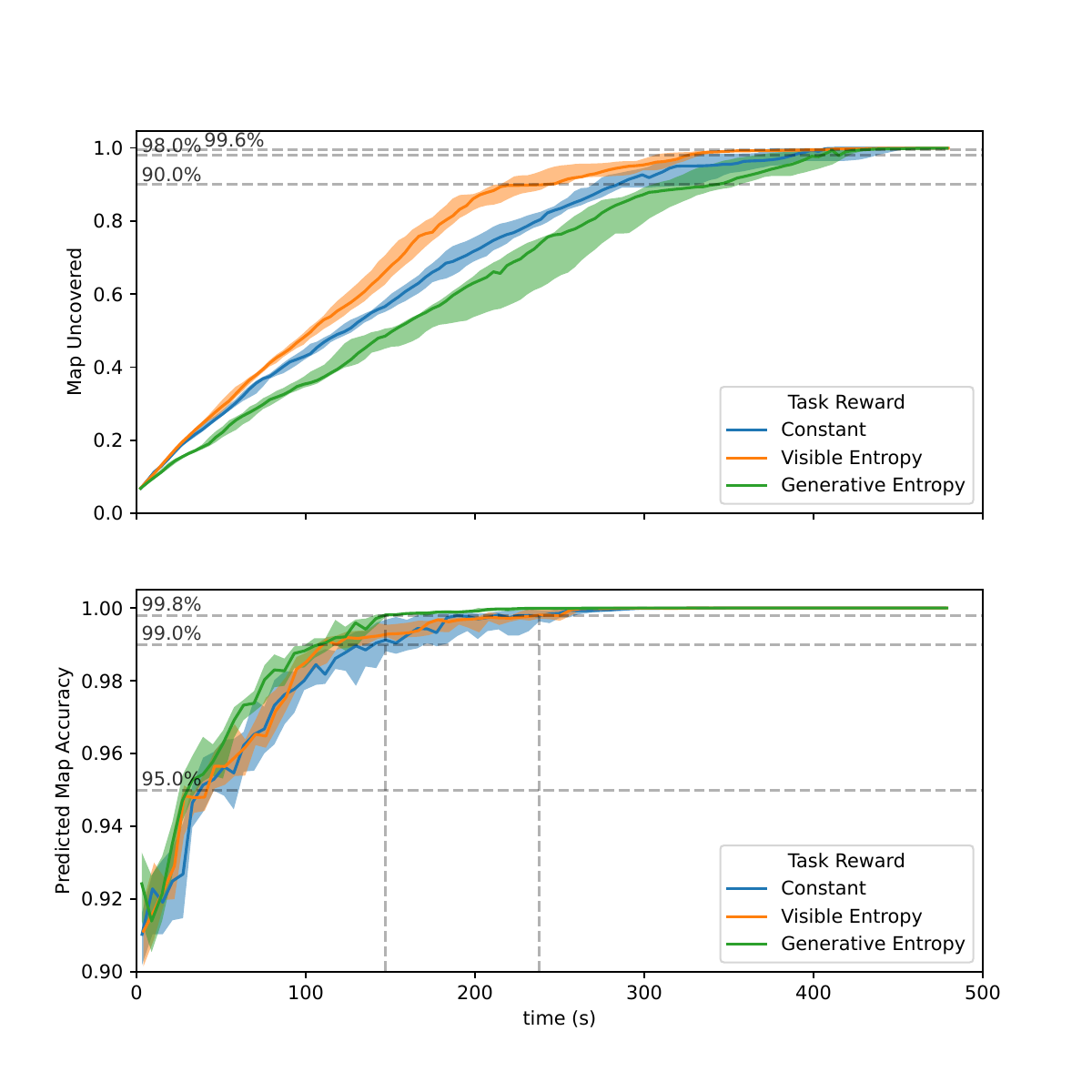}
\caption{Portion of map explored and accuracy of the predicted map over time, for three task reward settings. Each line shows the median performance across 10 trials, with the shaded area representing the interquartile range. Selected accuracy thresholds are indicated with dashed lines.
}
\label{fig:exploration_results}
\end{figure}

The results of the two performance metrics measured across the three task reward settings are shown in Figure~\ref{fig:exploration_results}.
We can see that prioritizing visible entropy has the best performance with regard to uncovering map cells.
The \emph{Generative Entropy} setting actually has worse performance compared to a constant reward setting in this respect.
However, we can see that prioritizing generative entropy improves the convergence time of the predicted map to the true map when compared with the other methods.

To discuss the accuracy of the predicted map, we select three thresholds (rough, medium, and high accuracy) that would be meaningful to a human user in the context of this dataset.
We can see that even at the start of a run with no information, the map predictor can achieve roughly 90\% accuracy.
This is because the environments are mostly empty space, and even gross mispredictions will correctly guess that the periphery of the map is empty, which accounts for a large portion of the volume.
We set the first threshold at about 95\% accuracy, where the predictions correctly match the rough layout of the town, perhaps missing a building or two. 
Even this first threshold begins to provide valuable intelligence about the layout of the scene.
At 99\% accuracy, the prediction will closely match the ground truth, having correctly predicted the location and general shape of each building, although some building edges and small features may not be fully resolved.
Finally, at 99.8\% accuracy, nearly all relevant geometric information is recovered in the map prediction.
The relative difference between subsequent threshold values represents a recovery of 80\% of the remaining information.

First, we can gauge the relative advantage of leveraging map prediction in terms of the time required to make useful information available to the user, regardless of the task allocation method.
Without map prediction, we can only report the data that was explicitly observed.
By counting unknown cells as a 50\% accurate guess, we can establish equivalent accuracy thresholds at 90\%, 98\%, and 99.6\% of cells uncovered, which are displayed in the top subfigure of Figure~\ref{fig:exploration_results}.
Even the best method for uncovering cells, \emph{Visible Entropy}, takes \SI{241}{s} on average to reach the first accuracy threshold. 
When using map prediction with \emph{Visible Entropy}, the predicted output reaches the rough accuracy threshold in \SI{42}{s}, and the more useful medium accuracy threshold in \SI{110}{s} on average.


Next, we will compare the relative performance of map prediction under the three task reward settings.
All three methods perform similarly during the early stages of exploration.
At this point, little is known about the environment, so any additional space uncovered improves the prediction quality.
The \emph{Generative Entropy} setting shows better performance throughout exploration, but the most significant advantage can be seen with respect to the high accuracy threshold. 
Towards the end of the exploration process, there tend to be small pockets of unexplored space near building edges and corners.
These pockets are not valued highly by the \emph{Visible Entropy} reward since they contain relatively few cells, but they are nevertheless necessary for resolving the last few details of the map geometry.
Using generative entropy allows the robots to correctly identify and prioritize these pockets before heading towards the map edges to clear the periphery.
On average, the \emph{Generative Entropy} reward achieves the high accuracy threshold at \SI{147}{s} (with an interquartile range of \SI{141}{s} to \SI{157}{s}), 62\% faster than \emph{Visible Entropy} at \SI{238}{s} (with an interquartile range of \SI{204}{s} to \SI{257}{s}).



\addtolength{\textheight}{-7.1cm}   

\section{Discussion and Conclusions}

In this work, we developed a multi-agent map prediction algorithm capable of predicting fixed-size urban scenes with any amount of observation data. 
We selected Stable Diffusion as the map prediction architecture due its requisite accuracy on the urban test set, prediction speed, and ability to condition on semantic information. Although we don't make use of the attention heads available in the model in this work, we expect it to become a valuable part of future work. We anticipate maps that can have subsections recursively inpainted in accordance with learned knowledge about the semantic structure of local areas.

When selecting a map prediction model, we observed that Stable Diffusion had the best immediate map foresight, possibly a result from a latent denoiser that was forced to capture more coarse features of the space of training maps than the other models. Notably, RePaint surpassed Stable Diffusion in performance towards the end of trials (although it takes much longer to run). This could motivate a hybrid prediction scheme where Stable Diffusion is run for the early stage of exploration with RePaint added later for final refinements to the predicted map. 

Additionally, LaMa had a tendency to produce artifacts on the periphery of the occupancy grid map. One possible source of these artifacts could be that the high frequency terms required for success of the Fast Fourier Convolution on the training set cause overfitting that results in outlier clutter in the test set. The architecture does not possess the image space embedding available in Stable Diffusion that may prevent the occurrence of artifacts.

Our results show that map prediction provides the expected value proposition of providing rough, medium, and high accuracy maps sooner than methods relying on explicit agent observation. 
We introduced a metric to guide exploration with predicted maps termed the ``generative entropy''. While there is ongoing discussion about responsible use of generative algorithms, we observe a way to leverage their predictive power without requiring conviction about any singular prediction. By prioritizing locations with high uncertainty in the predicted map, we prioritize ascertainment in regions of disagreement within the inpainter. We found that this led to faster convergence to the correct predicted map in our simulated experiments, attaining a high quality map estimate 62\% faster compared to information-theoretic task weighting. 

We find these results promising for continued investigation into the insight that map prediction and generative entropy provides for information-based exploration. Information maps might be amenable to a similar prediction scheme, and the concept of generative entropy may serve as a mechanism to apply these insights to objects such as scene graph predictors \cite{chen2019scene}.


\bibliographystyle{IEEEtran}
\bibliography{citations}

\end{document}